\pdfoutput=1
\documentclass[11pt]{article}
\usepackage{acl}
\usepackage{times}
\usepackage{latexsym}
\usepackage[T1]{fontenc}
\usepackage[utf8]{inputenc}
\usepackage{microtype}
\usepackage{inconsolata}
\usepackage[normalem]{ulem}

\usepackage{amsmath}
\usepackage{pifont}
\usepackage{amssymb}
\usepackage{bbm}
\usepackage{graphicx}
\newcommand*{\red}{\textcolor{red}}

\definecolor{dgreen}{rgb}{0.01, 0.75, 0.24}
\definecolor{saffron}{rgb}{0.96, 0.77, 0.19}
\newcommand{\cmark}{\text{\ding{51}}}
\newcommand{\xmark}{\text{\ding{55}}}
\newcommand{\cf}[1]{\tiny (#1)}

\DeclareMathOperator*{\argmax}{arg\,max}

\title{Model Editing by Standard Fine-Tuning}

\author{Govind Gangadhar \and Karl Stratos \\
  Department of Computer Science \\
  Rutgers University \\
  \texttt{\{govind.gangadhar, karl.stratos\}@rutgers.edu}
}

\begin{document}
\maketitle

\begin{abstract}
Standard fine-tuning is considered not as effective as specialized methods for model editing due to its comparatively poor performance.
However, it is simple, agnostic to the architectural details of the model being edited,
and able to leverage advances in standard training techniques with no additional work (e.g., black-box PEFT for computational efficiency),
making it an appealing choice for a model editor.
In this work, we show that standard fine-tuning alone can yield competitive model editing performance with two minor modifications.
First, we optimize the conditional likelihood rather than the full likelihood.
Second, in addition to the typical practice of training on randomly paraphrased edit prompts to encourage generalization, we also train on random or similar unedited facts to encourage locality.
Our experiments on the ZsRE and \textsc{CounterFact} datasets demonstrate that these simple modifications
allow standard fine-tuning to match or outperform highly specialized editors in terms of edit score.
\end{abstract}

\vspace{-1mm}
\section{Introduction}
\vspace{-1mm}

Model editing is a promising approach to combating incorrect or otherwise unwanted knowledge in LLMs.
Given a set of edits that assert desired information, the approach aims to alter the model so that it not only succeeds in memorizing
these edits (efficacy) but also applies the asserted information
to new prompts (generalization), \emph{without} changing inferences that should remain unchanged (locality).
There is clearly a trade-off between these metrics.
At one extreme, the model can achieve high efficacy by
memorizing the edits, but it will fail in generalization and locality.
At another extreme, the model can achieve high locality by
staying unmodified, but it will fail in efficacy and generalization.

Naively fine-tuning the model on the requested edits
is well known to perform poorly, especially in locality.
This motivated researchers to develop highly specialized model editors \citep[\textit{inter alia}]{mitchell2022fast,meng2022locating,meng2023massediting,hartvigsen2023aging,li2024swea}.
A central theme of these methods is ``minimally invasive editing'' achieved by
a careful selection of layers, model-specific adapter designs, and low-rank updates.
While technically interesting, they require a suite of assumptions
which may not be satisfied in other contexts (e.g., with different model architectures).
In contrast, standard fine-tuning is simple, completely agnostic to the details of the model,
and can take advantage of advances in standard training such as
parameter-efficient fine-tuning (PEFT) techniques for immediate computational efficiency.\footnote{We emphasize that our use of PEFT is only for computational convenience (i.e., we use black-box PEFT and are not concerned with its details such as LoRA vs others) and should not be confused with works that develop specialized adapters for model editing such as \citet{yu2024melo}.
}

\begin{table}[t!]
  {\small
  \centering
\setlength{\tabcolsep}{0.3em} 
{\renewcommand{\arraystretch}{1.2}
\begin{tabular}{lcccc}
\hline
Editor &  \begin{tabular}{@{}c@{}}Standard \\ model?\end{tabular}  & \begin{tabular}{@{}c@{}}Batched \\ edits?\end{tabular} &  \begin{tabular}{@{}c@{}}No extra\\ training?\end{tabular}   & \begin{tabular}{@{}c@{}}Effective \\ edits?\end{tabular}\\
\hline
Naive fine-tuning   & \textcolor{dgreen}{\cmark} & \textcolor{dgreen}{\cmark} & \textcolor{dgreen}{\cmark} & \red{\xmark}\\
MEND & \red{\xmark} & \textcolor{dgreen}{\cmark} & \red{\xmark} & \textcolor{dgreen}{\cmark}\\
ROME & \red{\xmark} & \red{\xmark} & \textcolor{dgreen}{\cmark} & \textcolor{dgreen}{\cmark}\\
MEMIT & \red{\xmark} & \textcolor{dgreen}{\cmark} & \textcolor{dgreen}{\cmark} & \textcolor{dgreen}{\cmark}\\
\hline
Our fine-tuning  &  \textcolor{dgreen}{\cmark} & \textcolor{dgreen}{\cmark} & \red{\xmark} & \textcolor{dgreen}{\cmark} \\
\hline
\end{tabular}
}
\caption{Conceptual comparisons of model editors.}
\label{tab:comparisons}
}
\vspace{-4mm}
\end{table}

In this work, we show that standard fine-tuning can yield competitive model editing performance, focusing primarily on improving the edit score.
We shift the focus from models and algorithms to
training objectives and data augmentation.
See Table~\ref{tab:comparisons} for conceptual comparisons between our approach and other editors.
Our fine-tuning uses two small but impactful modifications.
First, in line with the theme of minimal editing, we optimize the \emph{conditional}
likelihood (i.e., mask all tokens except the edited target).
Second, we augment the training data with random or similar unedited facts to encourage locality. This is in addition to the typical practice in existing works of training on randomly paraphrased edits to encourage generalization.
These simple modifications allow standard fine-tuning to match or outperform specialized editors in mass-editing, and also perform respectably in single-editing.

\vspace{-1mm}
\section{Task}
\vspace{0mm}

Our main task is mass-editing (i.e., performing multiple edits at the same time).
Let $\mathcal{V}$ denote the vocabulary.
A \textbf{fact} is a sentence $x \in \mathcal{V}^T$ that expresses
a subject-relation-object triple $(s,r,o)$ in natural language.
We follow the convention in the model editing literature and assume that
the object $o \in \mathcal{V}^m$ is the last $m$ tokens of $x$.
The prefix $\pi = (x_1 \ldots x_{m-1})$ expresses $(s,r)$ and is denoted as the \textbf{prompt}.
Let $\mathcal{E}$ denote a set of facts to enforce (i.e., requested edits).
Our goal is to edit a language model so that it upholds the relations expressed in $\mathcal{E}$
without changing its behavior on other facts.
In ZsRE, $\mathcal{E}$ consists of 10,000 factual statements
(e.g., ``The artwork Gideon's Way was by who? John Creasey'').
In \textsc{CounterFact}, $\mathcal{E}$ consists of 10,000 counterfactual statements (e.g., ``TextEdit, a product of Nintendo'').

An editor is evaluated by three competing metrics: efficacy, generalization, and locality.
Let $p_\theta$ denote a language model edited on $\mathcal{E}$.
In ZsRE, efficacy is the accuracy of $o = \argmax_y p_\theta(y|\pi)$ for $(\pi, o) \in \mathcal{E}$;
generalization is the accuracy of $o = \argmax_y p_\theta(y|\pi_{\text{\tiny par}})$
where $\pi_{\text{\tiny par}}$ is a paraphrase of $\pi$;
locality is the accuracy of $o_{\text{\tiny unrel}} = \argmax_y p_\theta(y|\pi_{\text{\tiny unrel}})$
where $(\pi_{\text{\tiny unrel}}, o_{\text{\tiny unrel}})$ is an unrelated fact.
In \textsc{CounterFact}, efficacy and generalization measure the accuracy of $p_\theta(o|\pi) > p_\theta(o_{\text{\tiny pre}}|\pi)$
and $p_\theta(o|\pi_{\text{\tiny par}}) > p_\theta(o_{\text{\tiny pre}}|\pi_{\text{\tiny par}})$
where $o_{\text{\tiny pre}}$ is a pre-edit object for $(\pi, o) \in \mathcal{E}$
(e.g., ``Apple'' in ``TextEdit, a product of Nintendo'').
Locality measures the accuracy of $p_\theta(o_{\text{\tiny pre}}|\pi_{\text{\tiny nb}}) > p_\theta(o|\pi_{\text{\tiny nb}})$
where $\pi_{\text{\tiny nb}}$ is a ``neighborhood'' unedited prompt whose target is $o_{\text{\tiny pre}}$
(e.g., ``Macintosh File System, a product of'' $\rightarrow$ ``Apple'').
In both datasets, the final \textbf{edit score} is the harmonic mean of
efficacy, generalization, and locality.

It is well known that naive fine-tuning, namely just optimizing

\vspace{-5mm}
\begin{align}
\min_\theta\; - \sum_{x \in \mathcal{E}}\; \log p_\theta(x) \label{eq:naive}
\end{align}
\vspace{-3mm}

\noindent
results in a poor edit score. The main reason is that while it
improves efficacy and possibly even generalization, it harms locality by changing the
model's predictions on unrelated or neighborhood prompts.
This led to the development of highly specialized editing methods,
with the implicit assumption that fine-tuning is not effective for model editing.

\begin{figure}
\begin{center}
{\footnotesize
\noindent\fbox{%
    \parbox{0.466\textwidth}{%
Fact: The mother tongue of Danielle Darrieux is English \\

Paraphrase augmentation (generated by the model): \\
1) \textit{The present invention relates.} The mother tongue of Danielle Darrieux is English\\

Random fact augmentation (from the training split): \\
1) Crate \& Barrel was founded in Chicago \\
2) Haines Borough is within Nevada
    }%
}
}
\end{center}
\vspace{-3mm}
\caption{An example from \textsc{CounterFact} along with paraphrase and random fact augmentation. 
}
\label{fig:augmented_data}
\vspace{-3mm}
\end{figure}

\vspace{-1mm}
\section{Method}
\label{sec:method}
\vspace{-1mm}

We propose a purely fine-tuning-based method that achieves a competitive edit score.
Our objective is a slight variation of \eqref{eq:naive}:

\vspace{-4mm}
\begin{align}
\min_\theta\; - \sum_{(\pi, o) \in \mathcal{E} \cup \mathcal{P} \cup \mathcal{R}}\;  \log p_\theta(o|\pi) \label{eq:ours}
\end{align}
\vspace{-3mm}

\noindent
There are two small but important differences between \eqref{eq:naive} and \eqref{eq:ours}.
First, we optimize the \emph{conditional} likelihood of the edit target $o|\pi$
rather than the full likelihood.
Our motivation is to make the training more focused (for efficacy and generalization) and minimize the damage caused by fine-tuning (for locality).

Second, we fine-tune not only on the requested edits $\mathcal{E}$ but
also additional facts in $\mathcal{P}$ and $\mathcal{R}$ to promote generalization and locality.
$\mathcal{P}$ is pseudo-paraphrases of the prompts in $\mathcal{E}$
obtained by prepending random words generated from the (unedited) model.
Paraphrase augmentation is universally used in existing model editors \citep{mitchell2022fast,meng2023massediting,li2024swea}.
Additionally, we augment with facts $\mathcal{R}$ that should not be altered by editing on $\mathcal{E}$.
There are many ways to obtain $\mathcal{R}$, but we find taking
\emph{random} facts from the training split to be simple and effective.
We filter $x \in \mathcal{R}$ to ensure that $x$ does not include
the subject-relation-object triple used in evaluation.
Figure~\ref{fig:augmented_data} shows an example from \textsc{CounterFact} (we only show 1 paraphrase and 2 random facts for illustration, in practice we use more).

We emphasize that the augmented sets $\mathcal{P}$ and $\mathcal{R}$
do not use the evaluation facts (otherwise it is trivial).
Given the requested edits $\mathcal{E}$, we perform the data augmentation
and fine-tune the model according to \eqref{eq:ours}.

\paragraph{Contrastive learning.} Given that our goal in \textsc{CounterFact} is to achieve
$p_\theta(o|\pi) > p_\theta(o_{\text{\tiny pre}}|\pi)$
where $o$ and $o_{\text{\tiny pre}}$ are the new and pre-edit objects for a given prompt $\pi$,
it is natural to consider a contrastive objective.
We experimented with DPO \citep{rafailov2023direct}
where we frame $o|\pi$ as the ``preferred'' response
over $o_{\text{\tiny pre}}|\pi$, optimizing the DPO loss
jointly with \eqref{eq:ours}.
However, we found that prompt masking and data augmentation are already
effective and do not benefit from contrastive learning.

\begin{table*}[th!]
{\small
\centering
\setlength{\tabcolsep}{0.3em} 
{\renewcommand{\arraystretch}{1.2}
\begin{tabular}{ll|cccc|cccc}
\hline
&         & \multicolumn{4}{c|}{ZsRE} &  \multicolumn{4}{c}{\textsc{CounterFact}} \\
&  Editor & Score & Efficacy & Generalization & Locality & Score & Efficacy & Generalization & Locality \\
\hline
&--- (original GPT-J) &  26.4 & 26.4 \cf{0.6} & 25.8 \cf{0.5} & 27.0 \cf{0.5} &  22.4  & 15.2 \cf{0.7} & 17.7 \cf{0.6} & 83.5 \cf{0.5} \\
\hline
\hline
& FT-W (21st layer w/ weight decay) & 42.1 & 69.6 \cf{0.6} & 64.8 \cf{0.6} &  24.1 \cf{0.6} & 67.6 & 99.4 \cf{0.1} & 77.0 \cf{0.7} & 46.9 \cf{0.6} \\
& MEND \citep{mitchell2022fast} & 20.0 & 19.4 \cf{0.5} & 18.6 \cf{0.5} &  22.4 \cf{0.5} & 23.1 & 15.7 \cf{0.7} & 18.5 \cf{0.7} & \textbf{83.0} \cf{0.5} \\
& ROME \citep{meng2022locating} & 2.6 & 21.0 \cf{0.7} & 19.6 \cf{0.7} &  0.9 \cf{0.1} & 50.3 & 50.2 \cf{1.0} & 50.4 \cf{0.8} & 50.2 \cf{0.6} \\
& MEMIT \citep{meng2023massediting} & 50.8 & 96.7 \cf{0.3} & 89.7 \cf{0.5} &  26.6 \cf{0.5} & 85.8 & 98.9 \cf{0.2} & 88.6 \cf{0.5} & 73.7 \cf{0.5} \\
& PMET \citep{li2024pmet} & 51.0 & 96.9 \cf{0.3} & 90.6 \cf{0.2} &  26.7 \cf{0.2} & 86.2 & 99.5 \cf{0.1} & 92.8 \cf{0.4} & 71.4 \cf{0.5} \\
\hline
  & FT                     & 44.8 & \textbf{99.9} \cf{0.0} & \textbf{98.9} \cf{0.2} & 21.4 \cf{0.5} & 52.8 & 79.6 \cf{0.8} & 58.5 \cf{0.8} & 36.8 \cf{0.7} \\
       & FT (21st layer)    & 42.9	& \textbf{99.9} \cf{0.0}	& 87.4 \cf{0.5} & 20.5 \cf{0.5} & 60.5 & 99.9 \cf{0.04} & 63.3 \cf{0.8} & 42.0 \cf{0.6} \\
       & FT + Mask          & 58.3 & 97.6 \cf{0.3} & 91.7 \cf{0.5} & 32.9 \cf{0.6} & 54.3 & 97.1 \cf{0.3} & 62.1 \cf{0.8} & 34.7 \cf{0.6} \\
       & FT + Mask + Para      & 56.1 & \textbf{99.9} \cf{0.0} & 98.7 \cf{0.2} & 29.9 \cf{0.5} & 63.7  & \textbf{100.0} \cf{0.0} & 92.5 \cf{0.4} & 38.0 \cf{0.6} \\
       & FT + Mask + Para + Rand  & \textbf{62.0} & \textbf{99.9} \cf{0.0} & 97.0 \cf{0.3} & \textbf{35.6} \cf{0.6} & \textbf{86.5} & 98.8 \cf{0.2} & \textbf{93.6} \cf{0.4} & 72.0 \cf{0.6} \\
       & FT + Mask + Para + Rand + DPO & --- & ---  & --- & --- & 85.5 & 98.8 \cf{0.2} & 93.4 \cf{0.4} & 70.1 \cf{0.6} \\
\hline
\end{tabular}
}
\caption{Mass-editing results on ZsRE and \textsc{CounterFact} (10,000 edits each) with GPT-J. The results of FT-W, MEND, ROME, and MEMIT are from \citet{meng2023massediting}. FT denotes naive fine-tuning on the requested edits; FT (21st layer) denotes FT only on the 21st layer. ``+ Mask'' means we mask the prompt. ``+ Para'' means paraphrase augmentation (which is involved in all the baselines). ``+ Rand'' means we augment the data with \emph{random} facts from the training split (not overlapping with any evaluation facts). ``+ DPO'' means we additionally optimize the DPO loss term using the changed target as the preferred response over the pre-edit target (only provided in \textsc{CounterFact}). Except for FT (21st layer), all our results use LoRA for computational efficiency.
}
\label{tab:main}
}
\vspace{0mm}
\end{table*}

\begin{table*}[th!]
{\small
\centering
\setlength{\tabcolsep}{0.3em} 
{\renewcommand{\arraystretch}{1.2}
\begin{tabular}{ll|cccc|cccc}
\hline
&         & \multicolumn{4}{c|}{ZsRE} &  \multicolumn{4}{c}{\textsc{CounterFact}} \\
&  Editor & Score & Efficacy & Generalization & Locality & Score & Efficacy & Generalization & Locality \\
\hline
&--- (original GPT-2 XL) &  22.5 &22.2 \cf{0.5} & 21.3 \cf{0.5} & 24.2 \cf{0.5} &  30.5  & 22.2 \cf{0.9} & 24.7 \cf{0.8} & 78.1 \cf{0.6} \\
\hline
\hline
& FT  & 45.9 & 99.6 \cf{0.1} & 82.1 \cf{0.1} &  23.2 \cf{0.5} & 65.1 & \textbf{100.0} \cf{0.0} & 87.9 \cf{0.6} & 40.4 \cf{0.7} \\
& FT-L & 40.1 & 92.3 \cf{0.4} & 47.2 \cf{0.7} &  23.4 \cf{0.5} & 66.9 & 99.1 \cf{0.2} & 48.7 \cf{1.0} & 70.3 \cf{0.7} \\
& KN & - & -  & - &  - & 35.6 & 28.7 \cf{1.0} & 28.0 \cf{0.9} & 72.9 \cf{0.7} \\
& KE & 41.8 & 65.5 \cf{0.6} & 61.4 \cf{0.6} &  24.9 \cf{0.5} & 52.2 & 84.3 \cf{0.8} & 75.4 \cf{0.8} & 30.9 \cf{0.7} \\
& MEND & 42.9 & 75.9 \cf{0.5} & 65.3 \cf{0.6} &  24.1 \cf{0.5} & 57.9 & 99.1 \cf{0.2} & 65.4 \cf{0.9} & 37.9 \cf{0.7} \\
& ROME & 47.9 & 99.8 \cf{0.0} & 88.1 \cf{0.5} &  24.2 \cf{0.5} & \textbf{89.2} & 100.0 \cf{0.1} & \textbf{96.4} \cf{0.3} & \textbf{75.4} \cf{0.7} \\
\hline
       & FT + Mask + Para + Sim  & \textbf{51.4} & \textbf{100.0} \cf{0.0} & \textbf{99.9} \cf{0.0} & \textbf{26.1} \cf{0.8} & 83.1 & 98.6 \cf{0.3} & 87.3 \cf{0.7} & 69.0 \cf{0.7} \\
\hline
\end{tabular}
}
\caption{Single-editing results on ZsRE (10,000 edits) and \textsc{CounterFact} (7,500 edits) with GPT-2 XL. ``+ Sim'' means we include similar facts (measured by Sentence-BERT) instead of random facts for locality supervision.
}
\label{tab:single_edit}
}
\vspace{-2mm}
\end{table*}

\vspace{-1mm}
\section{Related Work}
\vspace{-1mm}

We discuss existing model editors and the different assumptions they require to highlight the simplicity of our approach; we refer to \citet{zhang2024comprehensive} for an in-depth survey.

MEND \citep{mitchell2022fast} is a meta-learning method \citep{Sinitsin2020Editable,de2021editing} that predicts the change in the gradient.
Similar to our approach, it also involves an explicit training stage (on the training split of ZsRE).
It uses a special rank-one decomposition of the gradient for parameter efficiency.
In our case, we achieve parameter efficiency without special considerations
simply by leveraging black-box PEFT.

ROME \citep{meng2022locating} is a locate-then-edit method \citep{geva2021transformer,dai2022knowledge}
that first uses causal tracing to identify the feedforward layer to change, and then applies
a rank-one update to the layer's weight matrix.
While it does not involve an explicit training stage, it
does require an explicit knowledge of the subject $s$ in the edit
and its vector representation (which is heuristically induced).
It also relies on Wikipedia to estimate the covariance matrix of
the subject embeddings (needed for the rank-one update).
MEMIT \citep{meng2023massediting} extends ROME to multiple edits by carefully spreading updates to multiple layers.
Like ROME, it requires layer specification, subject embeddings,
and covariance statistics obtained from Wikipedia.

IKE \citep{zheng2023can} proposes in-context learning for model editing.
While similar to our approach in avoiding the need to develop a specialized editor,
it requires a strong model capable of effective in-context learning and is critically limited to single-editing.
Consequently, IKE focuses on single-editing experiments with large-scale LLMs, making their results not comparable to ours.

We define our scope as achieving a competitive edit score through standard fine-tuning.
In particular, we do not focus on preserving general capabilities of the model being edited.
Recent work shows that performance on a variety of downstream tasks drops to zero under any editor \citep{gu2024model}.
We leave preserving the general capabilities of an LLM with a fine-tuning editor as the next step of our work.

\vspace{-1mm}
\section{Experiments}
\vspace{0mm}

\subsection{Mass-Editing}
\vspace{0mm}

Our main results are on mass-editing with on ZsRE and \textsc{CounterFact}, following the same setting in  \citet{meng2023massediting}.
Each dataset provides 10,000 requested edits (i.e., $\mathcal{E}$).
As described in Section~\ref{sec:method}, we
augment the fine-tuning data with pseudo-paraphrases of the edit prompts $\mathcal{P}$ for generalization supervision and random facts $\mathcal{R}$
for locality supervision.
More specifically, for each edit in $\mathcal{E}$ we generate
15 paraphrases and take 20 facts from the training split
while ensuring that they do not contain any evaluation facts.
Thus the total number of facts we fine-tune on is 360,000.
We fine-tune GPT-J (6B) \citep{gpt-j} with LoRA \citep{hu2022lora} for computational efficiency.
We optimize \eqref{eq:ours}. The training takes around 2--2.5 hours on 8 GPUs.\footnote{The code is available at: \url{https://github.com/au-revoir/model-editing-ft}.}

We give the results in Table~\ref{tab:main}
where the row ``FT + Mask + Para + Rand'' corresponds to our final method.
We can make the following observations.
First, the proposed prompt masking in \eqref{eq:ours} (FT + Mask)
improves the performance of vanilla fine-tuning (FT).
In fact, on ZsRE this already outperforms MEMIT without any data augmentation.
Second, augmenting the data with paraphrased prompts substantially improves generalization (FT + Mask + Para).
Third, augmenting the data with random facts further improves locality
(FT + Mask + Para + Rand), allowing standard fine-tuning to match the
performance of MEMIT on \textsc{CounterFact}.
Adding the DPO loss on top of the final method does not yield improvements.

We perform additional experiments on the WikiRecent dataset \citep{cohen2023evaluating} with similar conclusions; details are given in Appendix~\ref{app:wikirecent}.

\vspace{-1mm}
\subsection{Single-Editing}
\vspace{0mm}

To compare with existing methods developed for single-editing (i.e., updating the model for a single fact),
we consider optimizing \eqref{eq:ours} one edit at a time.
We follow the setting in ROME and use GPT-2 XL (1.5B).
The training takes around 1.4 seconds per edit on average.
We give the results in Table~\ref{tab:single_edit}.
We find that it is helpful to include similar facts instead of random facts for locality
supervision, presumably because fine-tuning is more sensitive with small data size
thus requiring more care in the selection of demonstrations.
We use Sentence-BERT embeddings \citep{reimers2019sentence} and take 15 nearest facts as $\mathcal{R}$.
On ZsRE, our method outperforms all existing single-edit methods.
On \textsc{CounterFact}, our method lags behind ROME which is specifically optimized for single-editing,
but still achieves the second-best edit score.

\vspace{0mm}
\subsection{Generative Metrics}
\label{sec:gen}
\vspace{-1mm}

We report results on generative metrics in Table~\ref{tab:generative} (see Appendix~\ref{app:gen-single} for single-editing results).
Following prior work \citep{meng2022locating,meng2023massediting},
we report fluency (entropy of $n$-gram distributions) and
consistency (similarity score with a reference text).
Our fine-tuning methods take a hit on these metrics while improving the
edit score, showing that more work is needed to go beyond classification.
However, recent work shows that \emph{none} of the compared editors preserves
downstream performance \citep{gu2024model},
thus achieving this goal meaningfully is still an open problem.

We additionally show that we can easily incorporate considerations for generative performance into fine-tuning.
Specifically, we optimize $(1 - \gamma) L_1(\theta) + \gamma L_2(\theta)$ where $L_1(\theta)$ is the loss in \eqref{eq:ours} and $L_2(\theta) = -\sum_{w \in \mathcal{W}}\;  \log p_\theta(w)$
is a language modeling loss on random Wikipedia text. We choose Wikipedia articles
so that they do not overlap with those used for consistency evaluation. We set $\gamma = 0.1$. With this change (``+ Wikipedia Loss''), we obtain a significant improvement in both fluency and consistency at the cost of a modest drop in the edit score.

\begin{table}[t!]
{\small
\centering
\setlength{\tabcolsep}{0.3em} 
{\renewcommand{\arraystretch}{1.2}
\begin{tabular}{l|c|rr}
\hline
Editor & Score & Fluency & Consistency \\
\hline
--- (original GPT-J) &  22.4 &   622.4 \cf{0.3} & 29.4 \cf{0.2} \\
\hline
\hline
FT-W & 67.6 & 293.9 \cf{2.4} & 15.9 \cf{0.3} \\
MEND & 20.0 & 618.4 \cf{0.3} & 31.1 \cf{0.2} \\
ROME & 50.3 & 589.6 \cf{0.5} & 3.3 \cf{0.0} \\
MEMIT & 85.8 & 619.9 \cf{0.3} & 40.1 \cf{0.2} \\
PMET & 86.2& 620.0 \cf{0.3}& 40.6 \cf{0.2}\\
\hline
FT    & 52.8 & 626.1 \cf{0.4}	& 31.0 \cf{0.2} \\
FT + Mask  & 54.3 & 563.6 \cf{0.5} & 6.1 \cf{0.1} \\
FT + Mask + Para  & 63.7 & 550.7 \cf{0.6} & 4.7 \cf{0.1} \\
FT + Mask + Para + Rand & 86.5 & 352.0 \cf{1.5} & 5.2 \cf{0.2} \\
\hspace{4mm} + Wikipedia Loss & 84.8 & 609.2 \cf{0.6} & 29.2 \cf{0.2} \\
\hline
\end{tabular}
}
\caption{Fluency and consistency with mass-editing on \textsc{CounterFact}.
``+ Wikipedia Loss'' means we additionally optimize a language modeling loss
on Wikipedia text (Section~\ref{sec:gen}).}
\label{tab:generative}
}
\vspace{0mm}
\end{table}

\begin{table}[t!]
{\small
\centering
\setlength{\tabcolsep}{0.3em} 
{\renewcommand{\arraystretch}{1.2}
\begin{tabular}{l|cccc}
\hline
Editor & Score & Efficacy & Generalization & Locality \\
\hline
--- (GPT-2 XL) &  29.9 &  21.8 \cf{0.8} & 24.1 \cf{0.7} & 78.3 \cf{0.5} \\
\hline
\hline
ROME & 50.4 & 50.3 \cf{0.9}	& 49.4 \cf{0.8} & 51.6 \cf{0.6} \\
MEMIT & 71.5 & 79.9 \cf{0.7} & 66.2 \cf{0.8} & 69.8 \cf{0.5} \\
\hline
FT + M + P + R & 85.4 & 98.8 \cf{0.2} & 86.5 \cf{0.5} & 74.3 \cf{0.5} \\

\hspace{4mm} w/o LoRA & 86.9 & 98.8 \cf{0.2} & 91.2 \cf{0.4} & 74.4 \cf{0.5} \\
\hline
\end{tabular}
}
\caption{Mass-editing results on \textsc{CounterFact} with and without LoRA.}
\label{tab:massedit_counterfact_gpt2}
}
\vspace{-3mm}
\end{table}

\vspace{0mm}
\subsection{Analysis}
\vspace{0mm}

\paragraph{What is the effect of PEFT?}
We repeat mass-editing experiments with a smaller model (GPT-2 XL, 1.5B)
to compare the performance of LoRA vs full fine-tuning in Table~\ref{tab:massedit_counterfact_gpt2}.
We see that we achieve better generalization without LoRA, thus our improvement comes from conditional likelihood optimization and data augmentation, not the use of adapters
(i.e., unlike specialized model editors that are based on adapters).

\paragraph{Does layer selection help?}
Layer selection is an important component in many existing model editing works.
To see if it also benefits us, we fine-tune (without LoRA) only the layers 3--5 in GPT-J which are a subset of the layers chosen in MEMIT.
Table~\ref{tab:massedit_layer_specific} shows that we obtain a significant improvement compared to updating all layers (Table~\ref{tab:main}).
Thus layer selection helps on top of our already strong fine-tuning approach.

\paragraph{Subject knowledge vs data augmentation.}
A common but rather strong assumption in many state-of-the-art editors is that the subject phrase in a fact is known.
This knowledge critically allows the model to completely avoid changing at test time if the subject is not recognized,
giving a significant advantage in locality. For instance, with perfect subject knowledge, locality is optimal.
We apply our neighborhood fact augmentation to SWEA$\bigoplus$OS \citep{li2024swea} which holds a state-of-the-art result on \textsc{CounterFact} using subject knowledge (see Appendix~\ref{app:swea} for details).
Table~\ref{tab:massedit_swea_counterfact} shows that the augmentation does not help further. Hence subject knowledge obviates the need for data augmentation aimed at improving locality.

\begin{table}[t!]
{\small
\centering
\setlength{\tabcolsep}{0.3em} 
{\renewcommand{\arraystretch}{1.2}
\begin{tabular}{l|cccc}
\hline
Editor & Score & Efficacy & Generalization & Locality \\
\hline
FT + M & 63.0 &98.1 \cf{0.2} &61.2 \cf{0.8} & 47.4 \cf{0.6} \\

FT + M + P & 77.8 & 100 \cf{0.0} & 98.6 \cf{0.2} & 54.2 \cf{0.6} \\

FT + M + P + R & 91.1 & 98.8 \cf{0.2} & 96.8 \cf{0.3} & 80.2 \cf{0.5} \\
\hline
\end{tabular}
}
\caption{Mass-editing results on \textsc{CounterFact} with GPT-J but only fine-tuning layers 3--5.}
\label{tab:massedit_layer_specific}
}
\vspace{0mm}
\end{table}

\begin{table}[t!]
{\small
\centering
\setlength{\tabcolsep}{0.3em} 
{\renewcommand{\arraystretch}{1.2}
\begin{tabular}{l|cccc}
\hline
Editor & Score & Efficacy & Generalization & Locality \\
\hline
--- (GPT-J) &  22.4 &  15.2 \cf{0.7} & 17.7 \cf{0.6} & 83.5 \cf{0.5} \\
\hline
\hline

SWEA$\bigoplus$OS & 91.2 &99.5 \cf{0.1} &98.1 \cf{0.2} & 79.0 \cf{0.5} \\

SWEA$\bigoplus$OS + $\mathcal{R_E}$ & 91.2 & 99.5 \cf{0.1} & 97.7 \cf{0.2} & 79.4 \cf{0.5} \\
\hline
\end{tabular}
}
\caption{Mass-editing results (reproduced) on \textsc{CounterFact} by adding neighborhood augmentation to SWEA$\bigoplus$OS.}
\label{tab:massedit_swea_counterfact}
}
\vspace{-3mm}
\end{table}

\vspace{0mm}
\section{Conclusions}
\vspace{-1mm}

We have demonstrated that standard fine-tuning is sufficient to obtain strong edit performance
with a slight modification: optimizing the conditional likelihood and augmenting the data
with additional facts to promote locality.
Our results challenge the assumption that standard fine-tuning is ineffective as a model editor, suggesting that model editing could be achieved as part of standard training rather than through a specialized model editor.

\section*{Acknowledgements}

We thank the anonymous ARR reviewers for helpful feedback.

\section*{Limitations}

Our scope is limited to the classification aspect of model editing
rather than the generative aspect to demonstrate the viability of a fine-tuning editor.
Thus our approach is currently unable to preserve the LLM's general capabilities,
as is the case with existing mass editors.

\bibliography{custom}

\appendix

\begin{figure}[t!]
\begin{center}
{\footnotesize
\noindent\fbox{%
    \parbox{0.466\textwidth}{%
Paraphrase the given incomplete statement without changing the meaning. The same completion in the original input must also work for your paraphrase. Provide as many distinct paraphrases as you can come up with. \\

INPUT: The war during which Mario Stoppani was in the armed forces was \\
1) Paraphrased: The war Mario Stoppani was in the army in was \\
2) Paraphrased: Mario Stoppani was in the war called \\
3) Paraphrased: Mario Stoppani served in the army during \\

INPUT: The birth date of Luca Pianca is \\
1) Paraphrased: The date of birth of Luca Pianca is \\
2) Paraphrased: Luca Pianca was born in \\
3) Paraphrased: The day of Luca Pianca's birth is \\
4) Paraphrased: Luca Pianca's birth date is \\
5) Paraphrased: The birthday of Luca Pianca is \\

INPUT: The name of the architect of Ravenna Cathedral is \\
1) Paraphrased: Ravenna Cathedral was built by \\
2) Paraphrased: The person who built Ravenna Cathedral was \\
3) Paraphrased: The architect of Ravenna Cathedral is \\
4) Paraphrased: The architect behind the construction of Ravenna Cathedral is \\
5) Paraphrased: Ravenna Cathedral was built by the architect \\

INPUT: \{prompt\} \\
1) Paraphrased:
    }%
}
}
\end{center}
\caption{The few-shot prompt we use for paraphrase generation. We selected in-context examples from \textsc{CounterFact}.}
\label{fig:prompt}
\end{figure}

\section{Mass-Editing on WikiRecent}
\label{app:wikirecent}

To further validate our mass-editing results, we perform additional experiments on a simplified version of the WikiRecent dataset \citep{cohen2023evaluating}.
The original dataset consists of 570 and 1,266 facts from Wikidata for training and testing, where each fact has been recently modified (after July 2022).
An example prompt is $\pi = $``The name of the position held by Nicolaus Bergius is'' with the target object $o = $``bishop''.
Since the dataset does not provide paraphrase prompts for evaluating generalization, we create paraphrases of $\pi$ by few-shot prompting GPT (\texttt{gpt-3.5-turbo-0125}).
The prompt for GPT is given in Figure~\ref{fig:prompt}.
Since the task is relatively straightforward, we find the generated paraphrases preserve the original meaning (e.g., ``The position held by Nicolaus Bergius is'').
For neighborhood prompts, we select 10 random neighborhood prompts from \textsc{CounterFact} from the corresponding train/test splits.
The training data for our fine-tuning method consists of
1,266 edit requests, 18,990 augmented paraphrases (i.e., randomly generated), and 12,660 random neighborhood prompts from \textsc{CounterFact} (not overlapping with those in the test set).

\begin{table}[t!]
{\small
\centering
\setlength{\tabcolsep}{0.3em} 
{\renewcommand{\arraystretch}{1.2}
\begin{tabular}{l|cccc}
\hline
Editor & Score & Efficacy & Generalization & Locality \\
\hline
--- (GPT-J) &  37.4 &  34.4 \cf{1.7} & 34.5 \cf{1.5} & 45.3 \cf{0.9} \\
\hline
\hline
ROME & 35.0 & 39.8 \cf{2.2}	& 25.5 \cf{1.4} & \textbf{46.9} \cf{0.8} \\
MEMIT & 67.3 & 99.2 \cf{0.3} & 80.2 \cf{1.2} & 45.3 \cf{0.8} \\
\hline
FT & 55.8 & 68.1 \cf{1.8} & 60.4 \cf{1.7} & 44.5 \cf{0.8} \\
FT + M + P + R & \textbf{68.5} & \textbf{99.6} \cf{0.2} & \textbf{84.6} \cf{1.1} & 45.8 \cf{0.9} \\
\hline
\end{tabular}
}
\caption{Mass-editing results on WikiRecent (1,266 edits).}
\label{tab:wikirecent}
}
\vspace{-2mm}
\end{table}

Table~\ref{tab:wikirecent} shows the results.
We again find that while vanilla fine-tuning is not effective, our conditional fine-tuning with random data augmentation is competitive with MEMIT.

\section{Generative Metrics for Single-Editing}
\label{app:gen-single}

For completeness, we give generative performance for single-editing in Table~\ref{tab:generative-single}
as an accompaniment to Table~\ref{tab:generative}.
Our method is again competitive with the best performing single-editing baseline (ROME) in the edit score, but
has lower generative scores.

\begin{table}[t!]
{\small
\centering
\setlength{\tabcolsep}{0.3em} 
{\renewcommand{\arraystretch}{1.2}
\begin{tabular}{l|c|rr}
\hline
Editor & Score & Fluency & Consistency \\
\hline
--- (original GPT-2 XL) &  30.5 &  626.6 \cf{0.3} & 31.9 \cf{0.2} \\
\hline
\hline
FT & 65.1 & 607.1 \cf{1.1} & 40.5 \cf{0.3} \\
FT-L & 66.9 & 621.4 \cf{1.0} & 37.4 \cf{0.3} \\
KN & 35.6 & 570.4 \cf{2.3} & 30.3 \cf{0.3} \\
KE & 52.2 & 586.6 \cf{2.1} & 31.2 \cf{0.3} \\
MEND & 57.9 & 624.2 \cf{0.4} & 34.8 \cf{0.3} \\
ROME & 89.2 & 621.9 \cf{0.5} & 41.9 \cf{0.3} \\
\hline
FT + Mask + Para + Sim & 83.1 & 557.0 \cf{1.8} & 15.2 \cf{0.3} \\
\hline
\end{tabular}
}
\caption{Fluency and consistency with single-editing on \textsc{CounterFact}.}
\label{tab:generative-single}
}
\vspace{-2mm}
\end{table}

\section{Data Augmentation for SWEA$\bigoplus$OS}
\label{app:swea}

We train  SWEA$\bigoplus$OS on a set of random facts $\mathcal{R_E}$ in addition to random paraphrases (which is already present).
We use 10,661 random facts containing unique subjects from $\mathcal{E}$ within the training split.
This is done because SWEA$\bigoplus$OS is only concerned with a final cache containing subject tokens and their embeddings.
We first obtain the subject embeddings of the requested edits in $\mathcal{E}$ as described in \citet{li2024swea}.
We then repeat the process for the augmented random facts in $\mathcal{R_E}$.
Between the two sets of subjects, there are 392 subjects that overlap.
In the case of an overlap, we only retain the embeddings from the $\mathcal{E}$ and remove those from $\mathcal{R_E}$.

\end{document}